# Study of Constrained Network Structures for WGANs on Numeric Data Generation

**Wei Wang[1,2], Chuang Wang[1], Tao Cui[3], Yue Li[1,2]**
[1]College of Computer Science, Nankai University, Tianjin 300350, China
[2] Key Laboratory for Medical Data Analysis and Statistical Research of Tianjin (KLMDASR), Tianjin 300350, China
[3] NCMIS, LSEC, Academy of Mathematics and Systems Science, Chinese Academy of Sciences, Beijing 100190, China

Corresponding author: Yue Li (e-mail: liyue80@nankai.edu.cn).

**ABSTRACT** Some recent studies have suggested using GANs for numeric data generation such as to generate data for completing the imbalanced numeric data. Considering the significant difference between the dimensions of the numeric data and images, as well as the strong correlations between features of numeric data, the conventional GANs normally face an overfitting problem, consequently leads to an ill-conditioning problem in generating numeric and structured data. This paper studies the constrained network structures between generator G and discriminator D in WGAN, designs several structures including isomorphic, mirror and self-symmetric structures. We evaluates the performances of the constrained WGANs in data augmentations, taking the non-constrained GANs and WGANs as the baselines. Experiments prove the constrained structures have been improved in 17/20 groups of experiments. In twenty experiments on four UCI Machine Learning Repository datasets, Australian Credit Approval data, German Credit data, Pima Indians Diabetes data and SPECT heart data facing five conventional classifiers. Especially, Isomorphic WGAN is the best in 15/20 experiments. Finally, we theoretically proves that the effectiveness of constrained structures by the directed graphic model (DGM) analysis.

## I. INTRODUCTION

At present, multiple Generative Adversarial Network (GAN) schemes [1] have achieved significant progress in generating images and enhanced the accuracy of the classifier, where some of the GANs can produce almost indistinguishable images from human visional examination. In recent two years, several GAN models have been proposed for numeric data generation, which aims to generate samples to improve detection rates form multiple classifiers on the credit card fraud dataset [2, 3] and the telecom fraud dataset [4]. However, compared to the conventional data generation methods, taken Synthetic Minority Over-Sampling Technique (SMOTE) [5] as an example, the GAN based methods have not exhibited many advantages [6].

TABLE 1
AUC FOR FOUR DATASETS CLASSIFIED BY RF

| Techniques | Australian Credit Approval | German Credit | Pima Indians Diabetes | SPECT Heart |
|---|---|---|---|---|
| | AUC | | | |
| SMOTE | 0.9201 | 0.7600 | 0.8052 | 0.8092 |
| GAN | 0.9186 | 0.7359 | 0.7937 | 0.7634 |
| GAN-DAE | 0.9161 | 0.7360 | 0.8042 | 0.7598 |
| WGAN | 0.9147 | 0.7638 | 0.7755 | 0.7665 |

Table 1 demonstrates the AUC form classifier Random Forest (RF) as an example. It can be seen that the SMOTE generated data leads to a higher AUC under RF on four datasets [8]. The less effectiveness of GAN could be due to two factors, the lower dimensions on numeric datasets and the stronger correlation between the values on each dimension. Numeric data is usually low dimensional, such as Pima Indians Diabetes datasets with 8 dimensions and SPECT Heart datasets with 22 dimensions, in contrast to the image datasets, where the CIFAR-10 dataset with dimensions and MNIST dataset with dimensions. On the other hand, the values on each dimension generally represent a concrete meaning, such as Age in Pima Indians Diabetes dataset, whereas the values in an image are just a pixel with very little practical significance. Hence, the GAN-based data generation requests a much stronger generator for high-quality data generation.

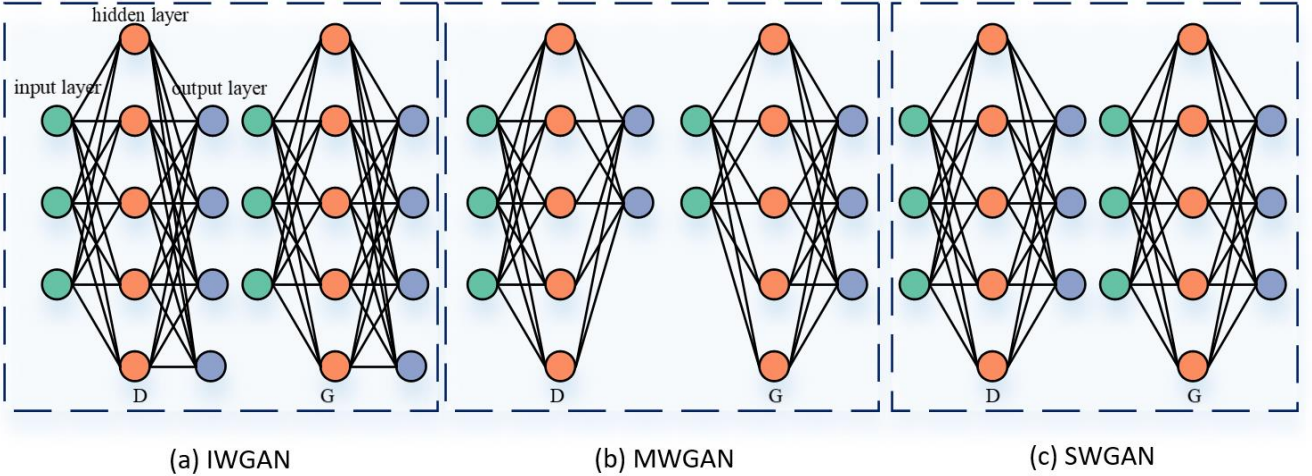

FIGURE 1. The Comparison of G-D pair's Structures of (a) IWGAN (b) MWGAN and (c) SWGAN

GANs normally face an overfitting problem, consequently leads to an ill-conditioning problem [7] in generating numeric and structured data. Beneficial from the Wasserstein distance as the loss function, we technically setup several constrained network pairs. We design several constrained network structures between G and D in WGAN for data generation. The structures of isomorphic (IWGAN), mirror (MWGAN) and self-symmetric WGAN (SWGAN) are shown in Figure 1. We define the isomorphic, mirror and self-symmetric structure for the G and D pair. Here the isomorphic structure is defined as that the two networks have the same number of layers, each layer has the same number of nodes, and every two neighboring layers have the same connection. The mirror structure is defined as mirror symmetrical network structure. The self-symmetric structure is defined as symmetrical network structure. The DGM analysis theoretically proves that constrained network structures provide additional restrictions in learning G from D, and verse vice, respectively.

In evaluating of GAN-based generated data, we compared IWGAN, MWGAN, SWGAN to three other GANs: conventional Wasserstein Generative Adversarial Network (WGAN) [6], adapted GAN proposed in 2017 [3], and GAN-DAE in 2018 [4]. In addition, the most widely used oversampling method, SMOTE [5], and is also employed in the evaluation as the baseline of data generation. Experiments are carried out on four widely studied datasets [8] and five classifiers, including Artificial Neural Network (ANN), Support Vector Machine (SVM), k-Nearest Neighbor (KNN), Gradient Boosting Classifier (GBC) and RF. In the common metrics, AUC in four datasets on five classifiers compared with three other GANs, and the conventional SMOTE methods add up to 20 groups of experiments. Experiments prove all the constrained structures have been improved in 20 groups of experiments. Especially, IWGAN outperforms all others in 15/20 groups. And the convergence rate of IWGAN is increased, and the initial error of loss function is reduced. Subsequently, some isomorphic structures may exist even if G and D do not satisfy the requirements of the numbers of layers and the same layers. Technically, even though the number of nodes in each layer of networks are slightly different, they can still be approximately isomorphic. We prove this point through the experiment and the analysis in Section VI. Moreover, Relevant follow-up studies may inspire us create other forms of GAN. As discussed in section VI, even if the isomorphism is technically imperfect or may correspond to other mapping relationships, it is still partially valid. In view of this, we further study the isomorphism of GAN partial layers in other cases, to improve their performance on numeric data and even image generation.

The remainder of the paper is organized as follows. We introduce an overview of previous related works on GAN and data generation in Section II. We provide the proposed approach and the analysis through DGM in detail in Section III. We show the further improved performance of constrained WGAN than SMOTE and other GANs using five classifiers on four representative datasets in Section VI. Finally, Section 5 presents the conclusions and outlines possible directions for future research.

## II. RELATED WORKS

GAN is designed based on the idea of competition [11] the objective of the G is to confuse the D. The D aims to distinguish the instances coming from the G and the instances coming from the original dataset. GAN is mainly used in the field of images to enhance the accuracy of the classifier [1]. The conventional method is to generate new samples by using SMOTE. SMOTE has been applied widely to the data generation. [5] Recently, GAN has been used to generate samples to improve classifier performance in credit card fraud detection [2, 3]. Zheng et al. [4] adopted a deep denoising autoencoder to learn the complex probabilistic relationship among the input features effectively and employed adversarial learning that establishes a min-max game between a discriminator and a generator to accurately discriminate between positive samples and negative samples in the data distribution. Larsen et al. [12] presented an autoencoder that leverages learned representations to better measure similarities in data space. By combining a variational autoencoder with a generative adversarial network, it can use learned feature representations in the GAN discriminator as the basis for the VAE reconstruction objective. The current GAN methods cause the unsatisfactory performance improvement of the classifier on the numeric datasets. The less effectiveness of GAN could be due to two factors, the dimensional differences and the representations in each dimension, as discussed in the introduction section.

## III. PROPOSED METHOD

### A. GAN and WGAN

Generative adversarial network (GAN) consists of two models, a generator model defined as G and a discriminator model defined as D. The GAN base on the idea of competition. The objective of the G is to confuse the D. The objective of the D is to distinguish the instances coming from the G and the instances coming from the original dataset.

More detailed, G: Z → X where Z is the noise space of arbitrary dimension d Z that corresponds to a hyper parameter and X is the data space, aims to capture the data distribution. D: X → [0, 1], estimates the probability that a sample came from the data distribution rather than G. G and D compete in a two-player min-max game with value function:

$$\min_q \max_w E_{x \sim P_r}\left[\log D_W(x)\right] - E_{z \sim P_z}\left[\log D_W(g_q(z))\right] \quad (1)$$

And the model structure of GAN is shown in Figure 2 below.

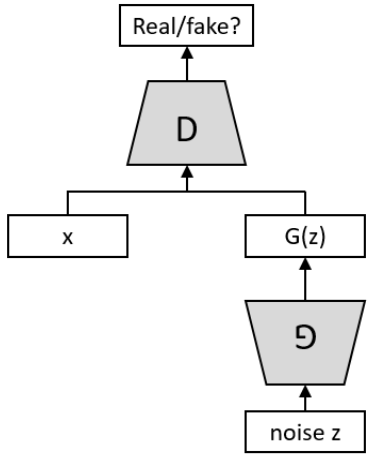

**FIGURE 2.** The model structure of GAN

However, the above methods have a series of problems such as collapse problem and the loss function of the generator does not converge. Wasserstein GAN (WGAN) proposed by Arjovsky M et al. [18] completely solved the problem of GAN training instability. WGAN uses Wasserstein distance (Earth-Mover):

$$W(P_r, P_\theta) = \inf_{\gamma \sim \prod(P_r, P_\theta)} E_{(x,y)}\left[\|x - y\|\right] \quad (2)$$

Where $\prod(P_r, P_\theta)$ denotes the set of all joint distributions $\gamma \sim (x, y)$ whose marginal are respectively $P_r$ and $P_\theta$.

Kantorovich-Rubinstein duality:

$$W(P_r, P_\theta) = \sup_{\|f\|_L \leq 1} E_{x \sim P_r}\left[f(x)\right] - E_{x \sim P_\theta}\left[f(x)\right]$$

$$= \frac{1}{K} \sup_{\|f\|_L \leq K} E_{x \sim P_r}\left[f(x)\right] - E_{x \sim P_\theta}\left[f(x)\right] \quad (3)$$

$$\Rightarrow \max_{\|f\|_L \leq 1} E_{x \sim P_r}\left[f(x)\right] - E_{x \sim P_\theta}\left[f(x)\right]$$

The loss function of WGAN shown as follows:

$$\min_\theta \max_\omega E_{x \sim P_r}\left[f_\omega(x)\right] - E_{z \sim P_z}\left[f_\omega(g_\theta(z))\right] \quad (4)$$

Compared to GAN, the problem of the collapse mode is almost solved by WGAN, ensuring the diversity of the generated samples.

### B. Constrained WGANs and Improvement

In order to design a stronger generator, we introduce constrained network structures between G and D so that a G-D pair constraint can be added in the respective learning of G and D. The challenge of setting up the constrained network structure is the dimension differences of the inputs and outputs of G and D, indicating different numbers of nodes in the first and last layer, between G and D as shown in Figure 1.

We employ the DGM to expound the working flow of WGAN and Constrained WGANs. It is noticeable that in the actual network learning, we do not adopt the conditional probability and transition probability from DGM to optimize the models. We apply DGM to describe the learning mechanism and further evaluate the effectiveness of constrained network structures in WGAN.

In DGM representation, we define G, D, E and $E_G$ as random variables in Hilbert space respectively. $E_G$ and E are error distributions under the influence of multiple random variables. Then the whole process of WGAN can be represented by DGM in Figure 3 (a). Moreover, for G and D learning, we define the function $f$ to represent the learning process for G and D gave the observed random variables

$$f(G | D, E_G, E) \text{ for learning } G \quad (5)$$

$$f(D | G, E_G, E) \text{ for learning } D \quad (6)$$

After factorization in DGM, there are

$$f(G|D, E_G, E) = f(G|E_G, D) \quad (7)$$

$$f(D|G, E_G, E) = f(D|E) \cdot f(D|E_G, G) \quad (8)$$

The above description conforms to the detail learning process of WGAN. When learning G via Equation (7), we calculate $E_G$ to minimize G under given D. In addition, when learning D from Equation (8), besides the constraints of E, we calculate to minimize D under given G, which satisfies the adversarial relationship between G and D.

At first, in the isomorphic structure, we define the constrained function that acts on the $G \rightarrow D$ Hilbert space given by:

$$D = TG \quad (9)$$

Correspondingly, in DGM we introduce a hidden variable T, and the discriminator function becomes the TG. It is noticed that we do not change the learning process of WGAN. So we still try to solve the optimal D=TG, not hidden variable T when learning stage for the discriminator.

By introducing T, we can see that two flows from $T \rightarrow D$ and $G \rightarrow D$ are added in DGM. Then the whole process of constrained WGAN can be represented by Figure 3 (b).

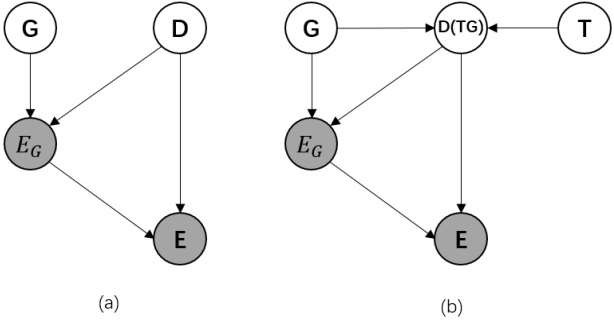

FIGURE 3. The processes of (a) WGAN and (b) Constrained WGANs represented by DGM

In the process of learning, the learning function becomes
$$f(G|T, TG, E_G, E) = f(G|E_G, TG) \cdot f(G|TG)$$
$$= f(G|E_G, D) \cdot f(G|D) \quad (10)$$
$$f(D|T, G, E_G, E) = f(TG|E) \cdot f(TG|E_G, G) \cdot f(TG|G)$$
$$= f(D|E) \cdot f(D|E_G, G) \cdot f(D|G) \quad (11)$$

Due to the existence of $T \rightarrow TG$, $f(G|D)$ for learning G and for learning D contain an extra constraint of $D \leftrightarrow G$, which are $f(G|D)$ in Equation (10) and $f(D|G)$ in Equation (11) added in the learning process of DGM. From a certain point of view, the mutual constraint of $D \leftrightarrow G$ explains why adding the constrained construction can generate stronger data for classifiers, improve the classification performances and guarantee a faster convergence speed.

### C. The Reason for Choosing the Constrained Structure

From DGM, it can be seen that if any hidden variable establishes the mapping of $G \rightarrow D$, restriction of $f(G|D)$ and $f(D|G)$ will be inserted in Equation (11) and Equation (12). Subsequently, there could be an improvement in the proposed method. In the alternative structures, constrained structures are selected for the following reasons.

a) Theoretically, the constrained function satisfies a one-to-one mapping from G space to D space. Assuming G function and D function follow some distributions in the two spaces, and then the constrained structures are the simplest structure for the transformation of these two distributions in theory. Other structures may not satisfy the one-to-one mapping (isomorphic, mirror and self-symmetric structures), so that the transformation probability distribution (i.e. $f(G|D)$ and $f(D|G)$) of $G \leftrightarrow D$ pairs are likely turned into N-P difficult problems without stable optimal solutions. Moreover, uncertainty G obviously leads to the instability of $P_x$ distribution, which determined by G. That is, an unstable output structure.

b) Technically, both WGAN and constrained WGAN do not employ DGM-based method to optimization for $(D, G)$. DGM is only used to discuss the association between G and D. In fact, $(D, G)$ is fitted separately by two neural networks. Then the simplest constrained structure between the two neural networks are isomorphic, mirror and self-symmetric structures.

Lastly, the same numbers of layers and the same layers are the sufficient and unnecessary condition for isomorphic relations. Some isomorphic functions may exist even if G and D do not satisfy the requirements of the numbers of layers and the same layers. Technically, even though the number of nodes in each layer of networks are slightly different, they can still be approximately isomorphic. We prove this point through the experiment and the analysis in Section V.

### V. EXPERIMENTS AND RESULTS

#### A. Experiments

The evaluation study is directed to determine the influence of the performance of constrained WGAN, GANs and SMOTE, using a variety of classifiers and two metrics, on four Datasets. According to this aim, experiments are conducted as follows:

*Datasets*: We conduct experimental analyses based on four datasets, which are obtained from the University of California Irvine (UCI) machine learning repository [8]. The four binary classification datasets with various imbalance ratios are Australian Credit Approval data, German Credit data, Pima Indians Diabetes data and SPECT heart data as shown in Table 2.

TABLE 2
DATASETS USED IN THE EXPERIMENTS

| Name of Dataset | No. of attributes | No. of attributes |
|---|---|---|
| Australian Credit Approval data | 690 | 14 |
| German Credit data | 1000 | 20 |
| Pima Indians Diabetes data | 768 | 8 |
| SPECT heart data | 267 | 22 |

***Classifiers***: Five general classifiers, Artificial Neural Network (ANN), Support Vector Machine (SVM), k-Nearest Neighbor (KNN), Random Forest Classifier (RF) and Gradient Boosting Classifier (GBC) are applied with the default parameter settings, recommended by scikit-learn [13].

***Evaluated Methods***: We compare the proposed IWGAN with the state-of-the-arts in augmenting data by GANs, including adapted GAN proposed in 2017 [3], GAN-DAE in 2018(Zheng, 2018), and the conventional WGAN under the baseline provided by SMOTE. Among these methods, WGAN without isomorphic structure request is used as another baseline for IWGAN.

***Evaluation metrics***: In this paper, AUC, as recommended in [10, 14, 15], are calculated as appropriate evaluation criterion to measure the data generation performance in classification. Partitions: We use 10-fold cross-validation in the experiments.

**B. Experiments**

***Evaluation on AUC***: Figure 4 demonstrates the AUC in four datasets on five classifiers compared with three other GANs, and the conventional SMOTE methods add up to 20 groups of experiments. Experiments prove the constrained WGANs have been improved in 17/20 groups of experiments compared with WGAN. And IWGAN outperforms all others in 15/20 groups. In the remaining five groups of experiments, the AUC index of IWGAN has three second best, two third best. Among the datasets, SPECT dataset is the least sensitive to the classifier and augmentation methods, while the other three datasets demonstrate the distinguishable result. About classifiers, GBC outputs the relatively best results on all of the datasets, while KNN produces the worsts. Affected by the datasets and classifiers, GAN-DAE and GAN generate unstable augmented data. For example, on all of the datasets classified by RF, the data which is generated by GAN-DAE and GAN has a worse result than data generated via SMOTE. However, when the datasets classified by SVM, GAN-DAE and GAN generated data illustrate better results than SMOTE in three of the four datasets. In our experiments, IWGAN outperforms all other GANs in all test cases except RF in Pima dataset. Meanwhile, IWGAN also outperforms the baseline, which is SOMTE, except RF in Pima dataset and SVM in SPECT dataset.

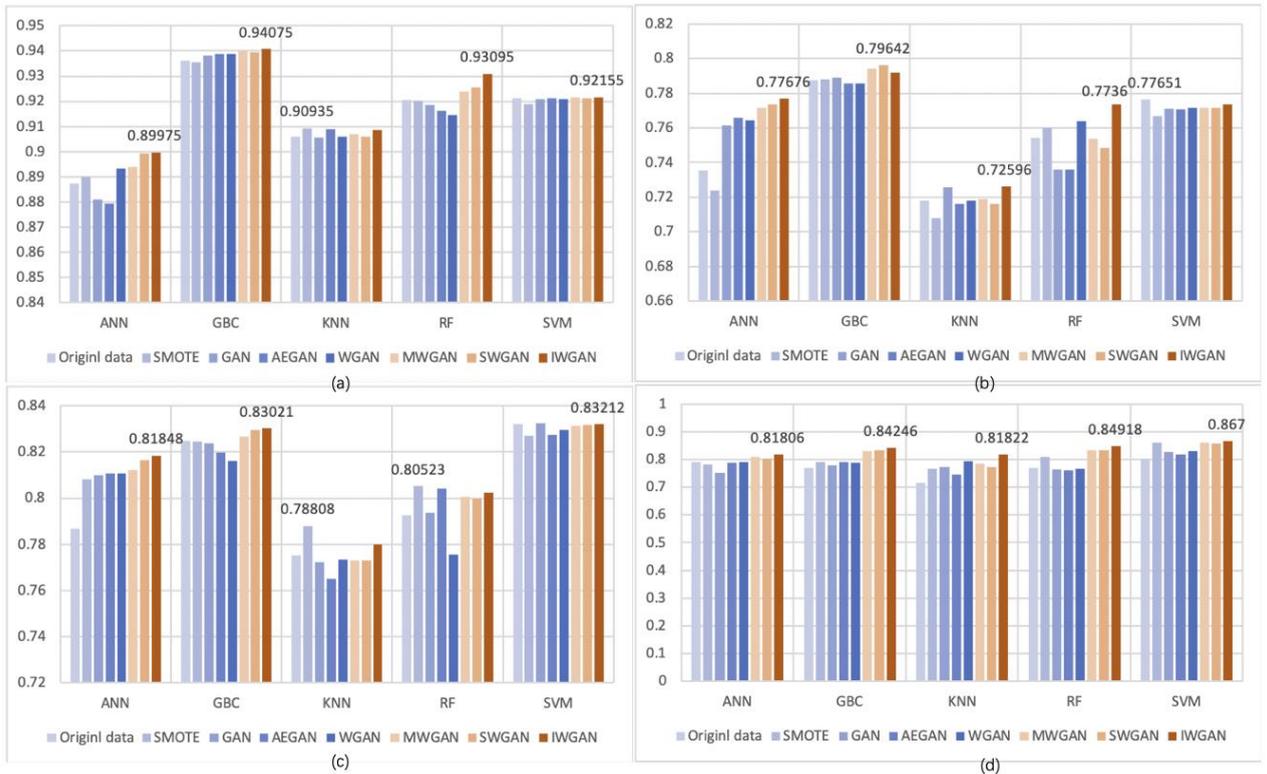

**FIGURE 4.** Five data augmentation methods of AUC for five classifiers in (a) Australian Credit Approval dataset (b) German Credit dataset (c) Pima Indians Diabetes dataset (d) SPECT Heart dataset

*Evolutions for Convergence Speed*: In the 10-fold cross-validation experiments, Occasionally, GAN and GAN-DAE do not converge, which is a widely discussed disadvantage [9, 11]. Therefore, in the discussion of convergence, we only compare the convergence of WGAN and IWGAN about generator G, as shown in Figure 5. It can be seen that IWGAN produces smaller initial loss function in Australian Credit Approval, German Credit and SPECT heart three datasets. Moreover, the initial error of IWGAN is only about 1/10 of WGAN. The initial error in Pima Indians Diabetes dataset is similar to that in WGAN. In terms of speed, the convergence rate of Pima Indians Diabetes datasets increases significantly. Moreover, the convergence speed of Australian Credit Approval and SPECT heart dataset improved slightly and were relatively stable. German Credit dataset's convergence rate has been slightly reduced. In general, on Australian Credit Approval dataset, Pima Indians Diabetes dataset and SPECT Heart dataset, the initial learning of G of IWGAN is close to the optimal global value. Another case on German Credit dataset is that although the initial G is not close enough to the optimal value, it can quickly approach the optimal value through constraints. In summary, IWGAN with isomorphic structure can enhance convergence performance.

These two types different performance on convergence may be because G-D pairs, $f(G|D)$ and $f(D|G)$, have different constraints performance explained by DGM. In one case, $f(G|D)$ results in larger optimization steps, which leads to fast convergence, but the loss function produces oscillation near the convergence point, as shown in Figure 6 (b). The other case is fast approaching the optimal solution although the convergence step is small. Both cases reflect the validity of $f(G|D)$ constraints in learning G. [17] However, we are still studying the network structure and $f(G|D)$ corresponding to its optimization step. It is noticeable that the blue line of WGAN in the Figure is lower, which means smaller $E_G$ loss, but not the optimal value and only the feedback error to D smaller. That is, it does not correspond to the better solution of this iteration [6]. From Equation (3), we can see that $E_G$ is the latter part, but the optimal solution corresponds to the global minimum distribution error, that is, the whole formula. In fact, our previous experiments have proved that IWGAN is more effective. In this experiment, we observe the convergence rate and stability.

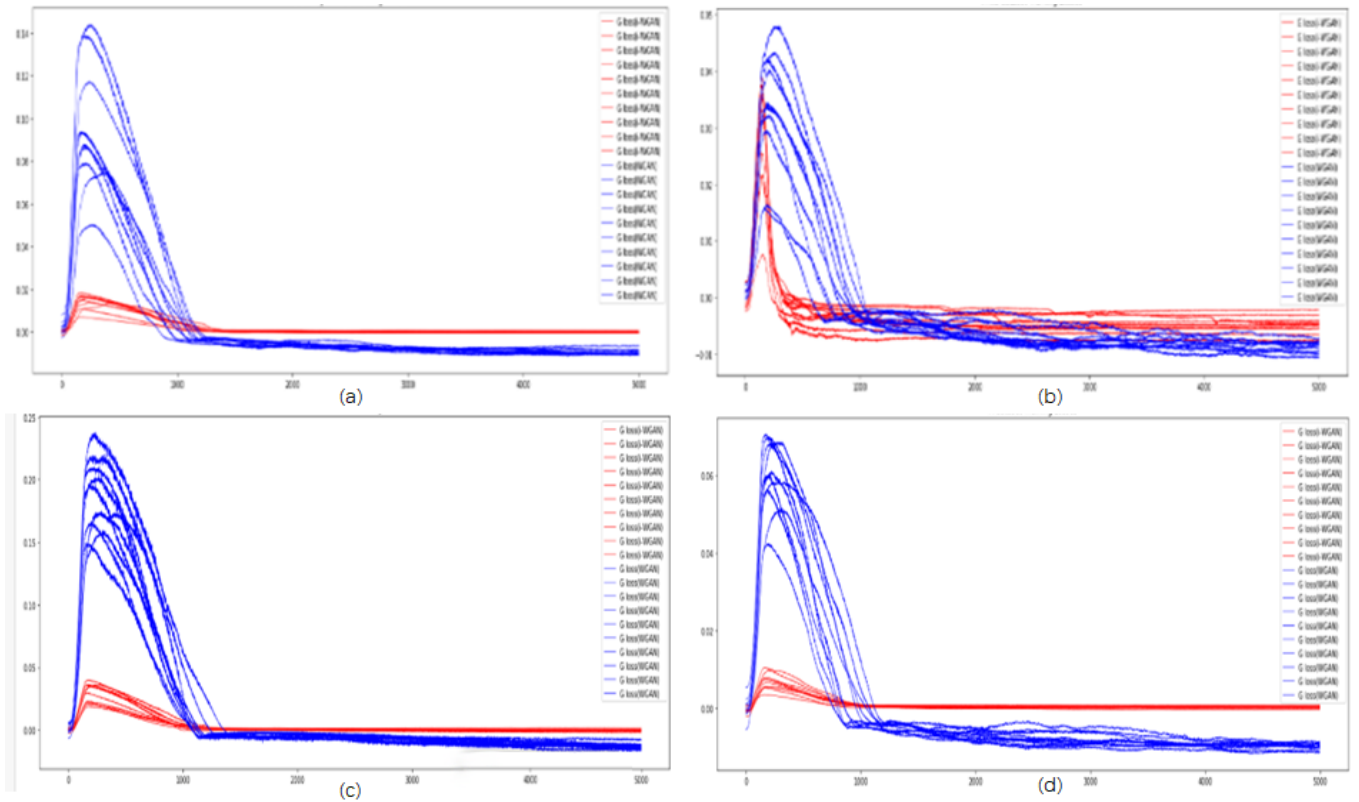

**FIGURE 5.** Comparison of the convergence of WGAN (blue lines) and IWGAN (red lines) about generator G in (a) German Credit dataset (b) Pima Indians Diabetes dataset (c) SPECT Heart dataset (d) Australian Credit Approval dataset

*Evaluation on Isomorphic Structure*: Some isomorphic functions may exist even if G and D do not satisfy the requirements of the numbers of layers and the same layers. We set up different IWGANs with relative isomorphic structures called r-IWGAN_1 to r-IWGAN_6, which have the same number of layers, but the number of D and G nodes is ±10%, ±20% and ±30% different.

In Figure 6, it can be indicated that the spots represent the AUC of each data augmentation algorithm under various classifiers in Germany credit dataset. Moreover, dotted lines represent the trend of different IWGANs. It can be seen that RF has an obvious trend, that is, the smaller the difference of the nodes' number in G and D, the higher the AUC of the RF classifier. Other three classifiers (KNN, GBC, ANN) have a certain trend. SVM has a relatively stable AUC value for different IWGANs. The reasons of SVM's stable performance is being explored. In conclusion, Experiments expose that the same layer and node can better represent the isomorphic structure. The experiment illustrates that the technical settings are not entirely isomorphic and can improve AUC on various classifiers. In other words, in Section III, if the theoretical function decomposition corresponds to the isomorphism of some sub-functions, it may also generate better data, which attracts our attention to design a more effective (G-D), to promote the performance on numeric data augmentation and even image generation in the future work.

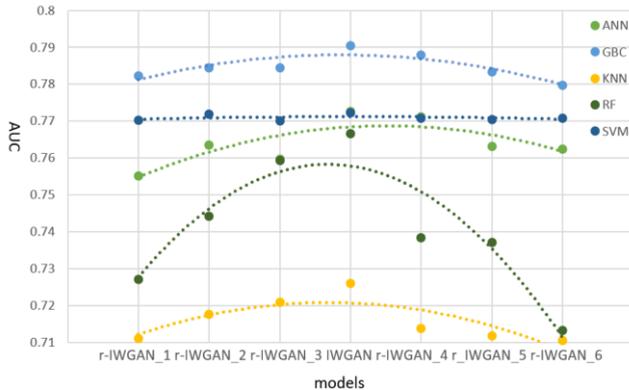

**FIGURE 6** Comparison of various relative IWGANs and IWGAN in Germany credit dataset

## VI. CONCLUSION AND FUTURE EXTENSIONS

This paper proposed IWGAN based on the isomorphic structure to data augmentation on four UCI publicly datasets. In addition, we technically setup the isomorphic network pairs, and the DGM analysis theoretically proves that this isomorphism provides an additional restriction in learning G from D, and verse vice, respectively. Moreover, in the common metrics, AUC in four datasets on five classifiers compared with three other GANs, and the conventional SMOTE methods add up to 20 groups of experiments. Experiments prove the constrained WGANs have been improved in 17/20 groups of experiments compared with WGAN. And IWGAN outperforms all others in 15/20 groups. In the remaining five groups of experiments, the AUC index of IWGAN has three second best, two third best. The convergence rate of IWGAN is increased, and the initial error of loss function is reduced. Subsequently, some isomorphic functions may exist even if G and D do not satisfy the requirements of the numbers of layers and the same layers. Technically, we set up different IWGANs with relative isomorphic structures, and find that the smaller the difference between the number of nodes in G and D, the higher the effect. If the theoretical function decomposition corresponds to the isomorphism of some sub-functions, it may also produce promotion, which attracts our attention to design a more effective G-D pair in future work. Relevant follow-up studies may inspire us to create other forms of GAN. Even if the isomorphism is technically imperfect or may correspond to other mapping relationships, it is still partially valid. Because of this, we further study the isomorphism of GAN partial layers in other cases, to improve their performance on numeric data augmentation and even image generation.

## ACKNOWLEDGMENT

This work was supported by the Tianjin Natural Science Foundations (17JCYBJC23000), National Key Research and Development Program of China (2016YFB0201304), Fundamental Research Funds for the Central Universities (Nankai University 070/63191114) and National Key Research and Development Program of China under the grant number (2018hjyzkfkt-002)